%% file: root.tex
\documentclass[letterpaper, 10 pt, conference]{ieeeconf}  

\IEEEoverridecommandlockouts                              

\usepackage{booktabs}
\usepackage{graphicx} 
\usepackage{algorithm}
\usepackage{algorithmicx}
\usepackage{algpseudocode}
\usepackage{caption}
\usepackage{subcaption}
\usepackage{multirow}
\usepackage{xspace}
\usepackage{wrapfig}
\usepackage{orcidlink}  
\usepackage{cleveref}

\newcommand{\sysname}{\texttt{Fast ECoT}\xspace}
\crefname{subsection}{Sec.}{secs.}
\crefname{figure}{Fig.}{secs.}
\crefname{algorithm}{Alg.}{secs.}
\crefname{table}{Tab.}{secs.}

\title{\LARGE \bf
Fast ECoT: Efficient Embodied Chain-of-Thought via Thoughts Reuse
}

\author{Zhekai Duan$^{1}$, Yuan Zhang$^{2}$, Shikai Geng$^{1}$, 
    Gaowen Liu$^{3}$, Joschka Boedecker$^{2}$, Chris Xiaoxuan Lu$^{*1}$ 
\thanks{$^{*}$ Corresponding author. Email: xiaoxuan.lu@ucl.ac.uk}
\thanks{$^{1}$ Department of Computer Science, University College London, UK }%
\thanks{$^{2}$ Department of Computer Science, University of Freiburg, Germany}%
\thanks{$^{3}$ Cisco Research, USA}        
}

\begin{document}

\maketitle
\thispagestyle{empty}
\pagestyle{empty}

\input{sections/abstract}
\input{sections/intro}
\input{sections/related}
\input{sections/background}
\input{sections/method}

\input{sections/experiments}

\input{sections/conclusion}

\bibliographystyle{IEEEtran} 
\bibliography{IEEEabrv, IEEEexample}

\end{document}

%% file: sections/abstract.tex
\begin{abstract}

Embodied Chain-of-Thought (ECoT) reasoning enhances vision-language-action (VLA) models by improving performance and interpretability through intermediate reasoning steps. However, its sequential autoregressive token generation introduces significant inference latency, limiting real-time deployment. We propose Fast ECoT, an inference-time acceleration method that exploits the structured and repetitive nature of ECoT to (1) cache and reuse high-level reasoning across timesteps and (2) parallelise the generation of modular reasoning steps. Additionally, we introduce an asynchronous scheduler that decouples reasoning from action decoding, further boosting responsiveness. Fast ECoT requires no model changes or additional training and easily integrates into existing VLA pipelines. Experiments in both simulation (LIBERO) and real-world robot tasks show up to a 7.5× reduction in latency with comparable or improved task success rate and reasoning faithfulness, bringing ECoT policies closer to practical real-time deployment. The code will be released upon acceptance.

\end{abstract}

%% file: sections/intro.tex
\section{INTRODUCTION}

Large-scale vision-language-action (VLA) models have recently advanced robotic control by leveraging internet-scale visual and textual knowledge~\cite{kim2024openvla, black2024pi0visionlanguageactionflowmodel, NVIDIAGR00T}. By combining pre-trained vision-language backbones with policy learning, these models exhibit impressive generalisation across open-world tasks. Among their most powerful capabilities is \textit{chain-of-thought} (CoT) reasoning—the ability to iteratively generate intermediate reasoning steps before taking an action. In robotics, \textit{Embodied Chain-of-Thought} (ECoT)~\cite{zawalski2024robotic} extends this concept by enabling robots to ``think out loud''—generating step-by-step textual reasoning traces (e.g., plans, subgoals, grounded visual inferences) at each time step, explicitly encoding the robot’s thought process before emitting an action. This augmentation improves model interpretability and boosts success rates.

While ECoT offers these benefits, they come at a steep computational cost. Generating reasoning traces involves producing dozens of tokens \textit{autoregressively} at each time step, resulting in sequential generation delays. As tasks grow more complex, the length of these reasoning chains increases, compounding the latency. In real-time robotic control, policies must react quickly to new observations; the inference overhead introduced by ECoT can slow the control loop to impractical speeds. In other words, the robot idles much of its time ``thinking'' rather than acting. Reducing this latency without sacrificing reasoning quality or task performance is essential for making ECoT viable in real-world deployments.

In this work, we address this inference bottleneck by proposing \sysname, a novel method for accelerating embodied chain-of-thought reasoning through thought reuse and parallelised reasoning. Our key insight is that ECoT outputs structurally and exhibits a high degree of temporal locality: many reasoning steps—such as recalling the task goal or rechecking the state of a target object—are repeated across time steps. Rather than regenerating the full reasoning trace at every step, we identify and cache recurring reasoning segments, reusing them in subsequent time steps. This reduces redundant computation while preserving the structure and interpretability of the thought process.

 Building on this reuse, we introduce a partial parallelisation strategy that transforms the traditionally sequential reasoning process into a batched one. By branching only when necessary to handle novel information, Fast ECoT enables multiple reasoning steps to be generated in parallel—substantially reducing inference latency. We further introduce an \textit{asynchronous scheduling} mechanism that decouples action and reasoning generation. Recognising that robot actions evolve faster than reasoning traces, we prioritise fast action decoding while allowing reasoning traces to update asynchronously in the background. This design reduces latency without compromising decision quality, as the cached reasoning remains stable over short time horizons.

%% file: sections/related.tex
\section{RELATED WORK}
\begin{figure*}[ht]
    \centering
    \includegraphics[width=\linewidth,trim=2cm 11cm 1.8cm 3.5cm, clip]{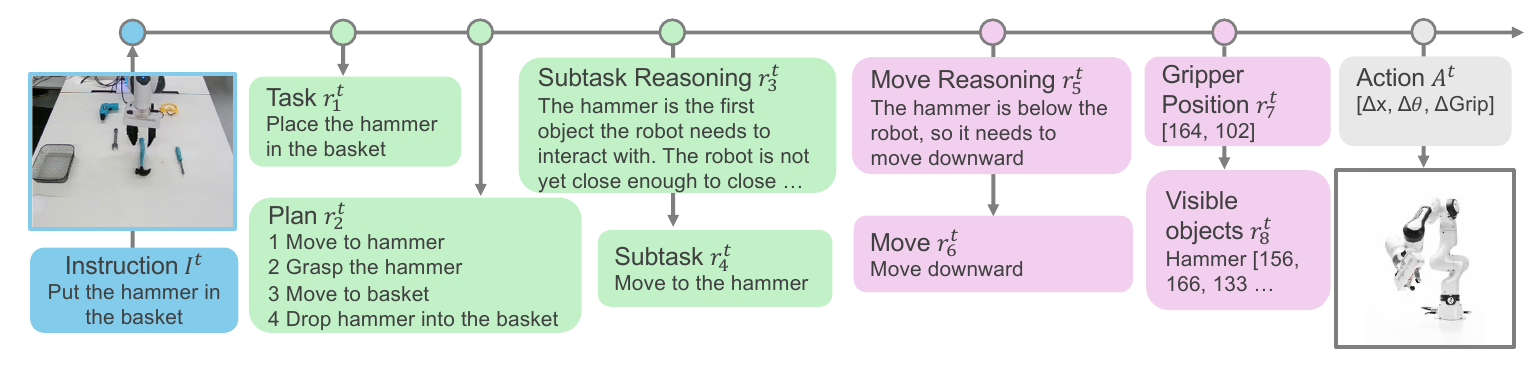}
    \caption{ECoT~\cite{zawalski2024robotic} reasoning autoregressively generates high-level \textbf{(green)} and low-level \textbf{(purple)} reasoning steps to enhance VLA performance.}
    \label{fig:ecot}
\end{figure*}

\textbf{Foundation Models for Robotic Manipulation.}  
Recent advances in robot learning have produced large-scale generalist policies that excel at robotic manipulation tasks, by first pre-training on diverse multimodal datasets~\cite{kim2024openvla, black2024pi0visionlanguageactionflowmodel, NVIDIAGR00T}, and further fine-tuning on extensive robot-specific data collections~\cite{walke2023bridgedata,embodimentcollaboration2024embodimentroboticlearning}.
Vision-language-action (VLA) architectures~\cite{kim2024openvla, black2024pi0visionlanguageactionflowmodel, NVIDIAGR00T}—which integrate vision-language models pretrained on internet-scale corpora~\cite{liu2023visualinstructiontuning, karamcheti2024prismaticvlmsinvestigatingdesign}—unify perception, language, and control into a single transformer-based policy, achieving state-of-the-art performance. 

Current VLA methods can be broadly categorized into two families: (1) \emph{monolithic models}, which integrate perception, language, and action within either single- or dual-system architectures~\cite{kim2024openvla, black2024pi0visionlanguageactionflowmodel, NVIDIAGR00T}; and (2) \emph{hierarchical models}, which explicitly decouple planning from policy execution by producing interpretable intermediate representations such as subtasks, keypoints, or programs~\cite{song2025acceleratingvisionlanguageactionmodelintegrated,li2025hamsterhierarchicalactionmodels}. 
While monolithic approaches highlight simplicity and end-to-end generalisation, hierarchical designs emphasise interpretability and modularity for long-horizon control.

\textbf{Reasoning for Robotic Control.}  
Chain-of-thought (CoT) prompting~\cite{wei2023chainofthoughtpromptingelicitsreasoning, ning2024skeletonofthoughtpromptingllmsefficient} has proven effective in enhancing large language models by encouraging step-by-step reasoning. 
Prior methods employ pre-trained language models for high-level planning~\cite{ahn2022can, zeng2022socraticmodelscomposingzeroshot, mees2023groundinglanguagevisualaffordances}, often requiring separate low-level controllers for execution. 
Recent work begins to integrate explicit reasoning end-to-end: ECoT introduces embodied chain-of-thought traces grounded in observations~\cite{zawalski2024robotic}, CoT-VLA predicts visual subgoal observations~\cite{zhao2025cotvla}, and RAD~\cite{RAD} and ThinkAct~\cite{thinkact} curate or align language reasoning for low-level action; however, these often rely on latent embeddings, generated sub-goals, or textual descriptions that are hard to precisely ground for manipulation. EMMAX~\cite{sun2024emmaxembodiedmultimodalaction} embeds reasoning as subtasks and predicted gripper states (2D/3D), but leverages limited full-scene context. In contrast, MolmoAct~\cite{lee2025molmoactactionreasoningmodels} performs “reasoning in space,” explicitly grounding each step in the scene so it can be decoded and visualised on the image plane and within the 3D environment, improving explainability and action prediction. Orthogonal to reasoning design, efficiency-oriented VLAs such as Spec-VLA~\cite{specvla} and FlashVLA~\cite{tan2025thinktwiceactonce} boost responsiveness by replacing strictly sequential decoding with speculative, parallel, or retraining-free acceleration. Our work builds on the ECoT but targets its chief bottleneck—autoregressive latency—via a partially parallelised reasoning framework that preserves interpretability while substantially reducing inference time.

\textbf{Inference Optimisation in Language and Multimodal Models.}  
A wide range of techniques has been proposed to speed up inference in autoregressive models. Speculative decoding~\cite{leviathan2023fastinferencetransformersspeculative, kim2023speculativedecodingbiglittle} accelerates generation by predicting tokens with a lightweight draft model and verifying them with the full model. Non-autoregressive and parallel decoding strategies have also been adopted, particularly in machine translation and, more recently, in robotic control~\cite{song2025acceleratingvisionlanguageactionmodelintegrated, kim2025finetuning}. Quantisation methods~\cite{lin2024awqactivationawareweightquantization, xiao2023smoothquant, huggingface} are widely used to reduce model precision for faster computation.
In contrast to these methods, which typically operate at the token or model level, our work focuses on \textit{reasoning-level acceleration}—breaking CoT generation into reusable, semantically meaningful segments that can be cached and executed in parallel. This approach is model-agnostic and does not rely on auxiliary models, fine-tuning, or precision reduction. By enabling efficient reuse and branching during reasoning, our method offers a lightweight and integrable solution for speeding up VLA policies with CoT reasoning, without compromising interpretability or task success. 


%% file: sections/background.tex
\section{PRELIMINARIES}
\subsection{Embodied Chain-of-Thought Reasoning (ECoT)}
Vision‑Language‑Action (VLA) models~\cite{kim2024openvla} build on large pre‑trained vision‑language models, fine‑tuning them to map a natural language instruction $I^t$ and image observation $O^t$ directly to low‑level robot actions $A^t$ via autoregressive token prediction at each control step $t$. Embodied Chain‑of‑Thought (ECoT)~\cite{zawalski2024robotic} reasoning augments this reactive paradigm by supervising the model to generate a structured sequence of $N$ intermediate reasoning steps $R^t = \{\, r_i^t \mid i = 1,2,\dots, N\}$, e.g. rephrased task, high‑level plan, grounded sub‑task, low‑level move command, visual features, before emitting the final action $A^t$. The reasoning steps are separated into high- and low-level steps (see \cref{fig:ecot}). 


\subsection{Continuous Batching}
In autoregressive models, batching improves throughput, but static batching can be inefficient when sequence lengths vary—shorter sequences must wait for longer ones, wasting compute on padding. Continuous batching \cite{280922} addresses this by dynamically replacing completed sequences with new ones, maintaining high GPU utilization and minimising idle time. This strategy, adopted by the work 
\cite{NvidiaTensorrt, kwon2023efficientmemorymanagementlarge}, has shown 2–4× throughput gains in LLM serving. 

%% file: sections/method.tex
\section{Method}

\begin{figure*}[!ht]
  \begin{subfigure}[t]{0.5\linewidth}
    \centering
    \includegraphics[width=\linewidth]{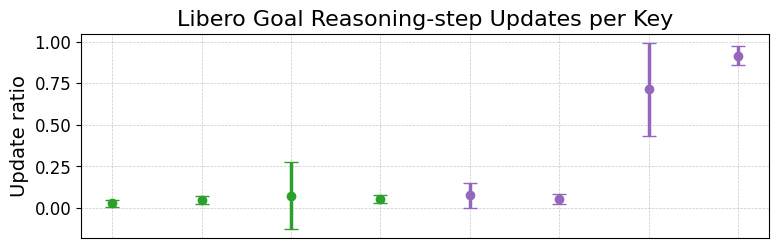}
    \label{fig:reasoning_update_libero}
  \end{subfigure}
    \hfill
  \begin{subfigure}[t]{0.5\linewidth}
    \centering
    \includegraphics[width=\linewidth]{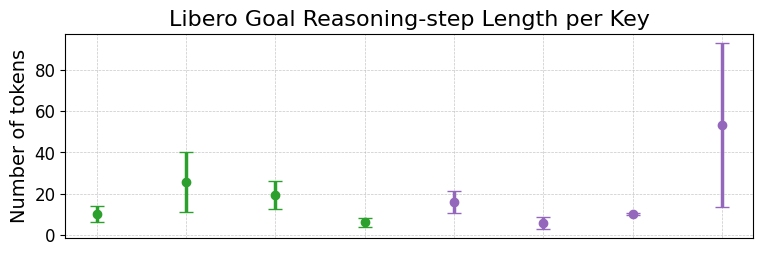}
    \label{fig:reasoning_length_libero}
  \end{subfigure}
  
  \vspace{-1mm}
  \centering
  \begin{subfigure}[t]{0.48\linewidth}
    \centering
    \includegraphics[width=\linewidth]{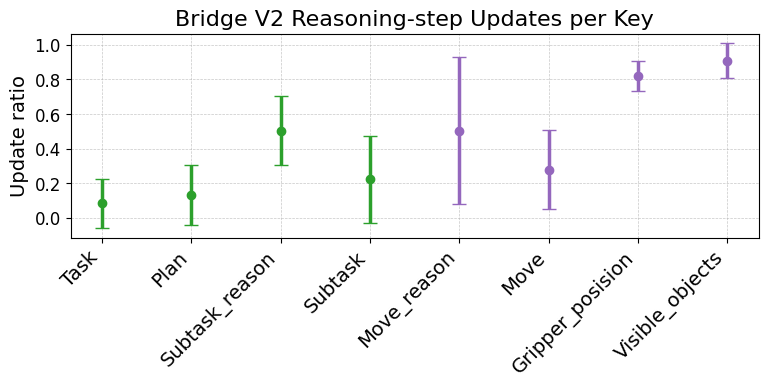}
    \caption{Mean update ratio ($\pm$1 standard deviation) per reasoning step, showing high-level stability versus frequent low-level updates.}
    \label{fig:reasoning_update}
  \end{subfigure}
  \hfill
  \begin{subfigure}[t]{0.48\linewidth}
    \centering
    \includegraphics[width=\linewidth]{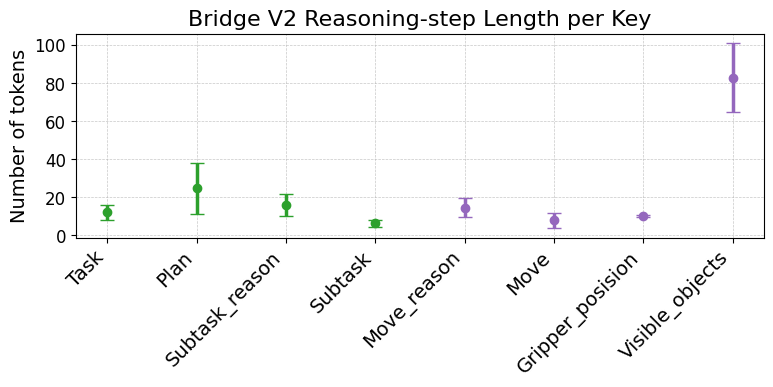}
    \caption{Mean length ($\pm$1 standard deviation) of different reasoning steps, showing significant disparities in token counts.}
    \label{fig:reasoning_length}
  \end{subfigure}
  \vspace{1ex}
  \caption{
    Statistics illustrating the pattern of ECoT reasoning steps under Libero Goal~\cite{liu2023libero} and Bridge V2 \cite{walke2023bridgedata}.
  }
  \label{fig:reasoning_stat}
\end{figure*}

\subsection{Inference Characteristics of ECoT Reasoning}
Unlike traditional chain-of-thought approaches in language modelling~\cite{wei2022chainofthought, deepseekai2025deepseekr1incentivizingreasoningcapability} and vision-language modelling~\cite{lu2023chameleonplugandplaycompositionalreasoning, xu2025llavacotletvisionlanguage}, which yield diverse, dynamic reasoning patterns that rarely \textit{recur}, ECoT consistently follows a structured, periodically recurring workflow: planning, sub-task identification, motion reasoning, and visual-feature processing, before predicting subsequent robot actions. To analyse this behaviour quantitatively, we sample all 
episodes containing reasoning from the Bridge V2 dataset~\cite{walke2023bridgedata} and compute the average and the standard deviation of the update ratio (percentage of reasoning content updated at the next time step) and token length for each reasoning step. As depicted in~\cref{fig:reasoning_update}, higher-level reasoning components in ECoT (e.g., planning and subtask reasoning) remain relatively similar across multiple time steps within an episode\footnote{While profiling the update ratios, we observed substantial variability in the visible detection module. This counterintuitive result stems from the instability inherent in the visual reasoning module of ECoT and the open-vocabulary nature of object labels, whereby identical objects may be assigned varying labels over time.}. For example, the planning module exhibits an average update ratio of only $8.4\%$, meaning $91.6\%$ of its reasoning content is unchanged and can be reused in subsequent inference steps without autoregressive regeneration. Leveraging this temporal locality, we propose to cache ECoT reasoning for reuse in successive time steps, thus potentially enabling the parallel generation of each reasoning step.
We observe a similar trend in the simulated environment, LIBERO-Goal~\cite{liu2023libero}: high-level reasoning updates rarely while low-level components refresh more frequently (see \cref{fig:reasoning_stat}); This consistency supports caching high-level reasoning and densely updating low-level content.


\begin{figure*}[!ht]
    \centering
    \includegraphics[width=1\linewidth, trim=0.8cm 11cm 9.7cm 4cm, clip]{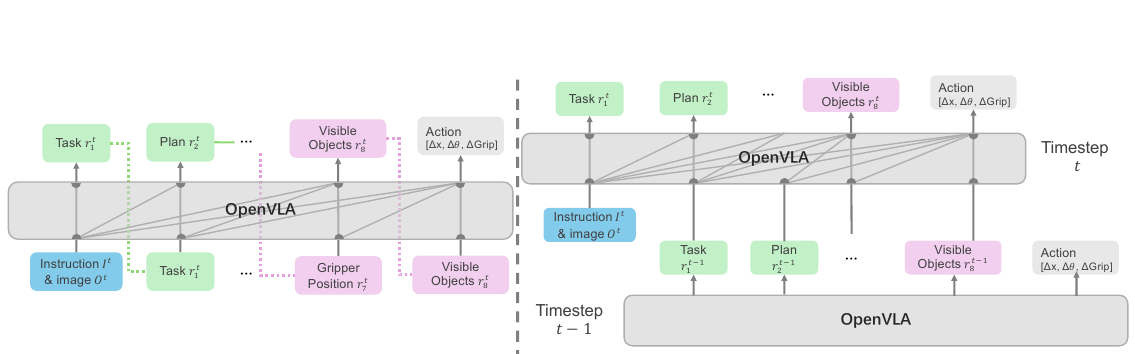}
    \caption{Comparison between ECoT \textbf{(left)} and our proposed \sysname \textbf{(right)}. Both decompose reasoning into fixed stages (e.g., task, plan, object grounding), but ECoT generates these sequentially at every step, while \sysname enables parallel generation and reuses cached higher-level reasoning from previous timesteps as context. The dotted lines coloured in green/magenta represent token copying.}
    \label{fig:batch_ecot}
\end{figure*}

\subsection{Parallelising Reasoning and Action Generation}

Compared to the original ECoT—which generates the full reasoning sequence sequentially and autoregressively at every timestep (see \cref{fig:batch_ecot} left)—our Fast ECoT reformulates each reasoning step \( r_n^t \) as a standalone generation task. For each step, we construct the input by combining the current observation \( O^t \), instruction \( I^t \), and the previously generated reasoning steps \( R^{t-1} = \{ r_i^{t-1} \mid i = 1,2,\dots,n-1 \} \) from the last timestep as prefix context. This allows all reasoning steps and the action at timestep \( t \) to be generated independently and in parallel, rather than waiting for preceding components to finish  (see~\cref{fig:batch_ecot} right).

\begin{figure}[!ht]
    \centering
    \includegraphics[width=0.9\linewidth,trim=4.6cm 11.7cm 6.6cm 3.5cm, clip]{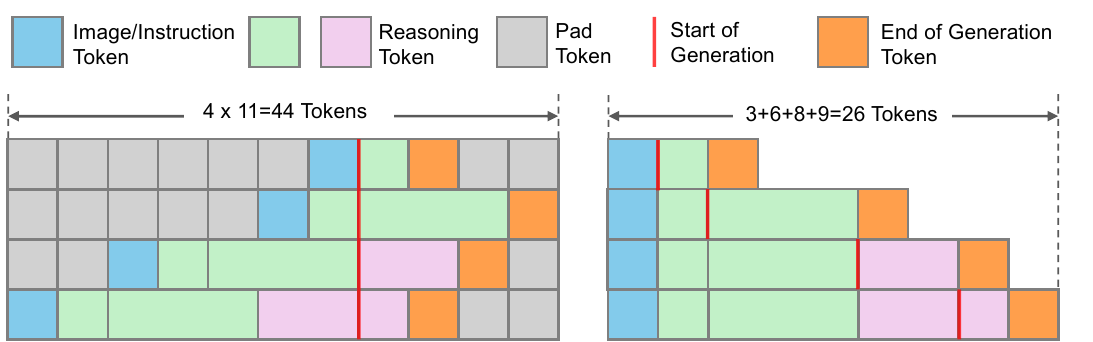}
    \caption{Illustratively comparing static vs. continuous batching for reasoning generation. \textbf{Left:} Static batching pads to the longest sequence, processing 4×11=44 tokens. \textbf{Right:} Continuous batching processes only actual tokens (3, 6, 8, 9), adding up to 26 tokens, which reduces padding and improves efficiency.}
    \label{fig:continuous_batching}
    \vspace{-2.5mm}
\end{figure}

While conceptually straightforward, this strategy introduces performance overhead. Since each reasoning module prepends a growing context of prior thoughts, the resulting input lengths vary significantly—early steps have short prompts, while later ones accumulate more history. Additionally, output lengths also differ across reasoning steps: low-level steps like gripper commands may require fewer than 20 tokens, while object-grounding reasoning can exceed 120 (see \cref{fig:reasoning_length}). Traditional static batching~\cite{280922} (see \cref{fig:continuous_batching}) handles such variability by padding all sequences in a batch to match the longest, which leads to substantial inefficiency on modern accelerators. This results in wasted compute on padding tokens, poor GPU utilisation, and offsets the gains from parallel generation.

To address inefficiencies from padding, we adopt continuous batching \cite{280922}, a dynamic scheduling strategy used in modern LLM serving engines~\cite{NvidiaTensorrt, kwon2023efficientmemorymanagementlarge}. Instead of padding all sequences to a fixed length, continuous batching maintains a queue where completed sequences are immediately replaced by new ones, allowing variable-length inputs to be processed efficiently. This minimises wasted computation on padding tokens, significantly improves GPU utilisation, and can reduce the total number of tokens processed by up to 6× (see \cref{fig:continuous_batching}). We use vLLM~\cite{kwon2023efficientmemorymanagementlarge} as our inference backend, and the pseudocode of \sysname is shown in \cref{alg:ecot_continuous_batched}.

\begin{algorithm}[!ht]
\caption{Parallel Embodied Chain‑of‑Thought}
\label{alg:ecot_continuous_batched}
\begin{algorithmic}[1]
\Require Time step $t$, Instruction $I^t$, Observation $O^t$, Last reasoning steps $R^{t-1}$
\State $c^t \gets \mathrm{encode}(I^t, O^t)$
\State $R^t \gets [\,]$
\For{$i = 1 \,\mathbf{to}\, N+1$ \textbf{concurrently}} 
  \State Autoregressively sample $r_i^t \sim P\bigl(r_i^t \mid c^t, R^{t-1}[:i]\bigr)$
\EndFor
\State \textbf{Synchronize}
\State $R^t,A^t \gets \mathrm{decode}(R^t[:-1],R^t[-1])$
\State \Return $R^t,A^t$
\end{algorithmic}
\end{algorithm}
    
\label{method:sync}

\vspace{-5pt}
\subsection{Asynchronous Reasoning and Action Updates}
\begin{figure}[ht]
    \centering
    \begin{subfigure}[t]{\linewidth}
        \centering
        \includegraphics[width=1.0\linewidth,trim=1cm 11cm 11.7cm 4cm, clip]{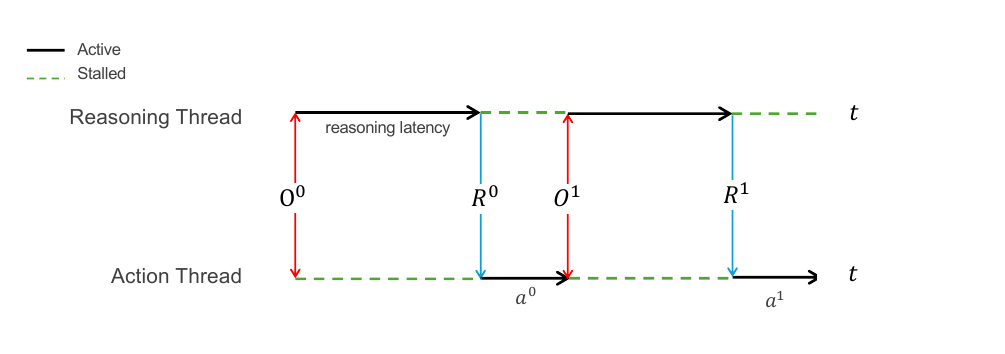}
        \caption{Synchronous inference.}
        \label{fig:flow_ecot_a}
    \end{subfigure}

    \begin{subfigure}[t]{\linewidth}
        \centering
        \includegraphics[width=1.0\linewidth,trim=1cm 5.5cm 11.7cm 10cm, clip]{figures/async_flow.pdf}
            \caption{Asynchronous inference.}
        \label{fig:flow_ecot_b}
    \end{subfigure}

    \caption{ Illustration of inference stacks. Asynchronous overlap reduces stall time (green dashed) of robot operation and increases action throughput within the same time window compared to synchronous execution.}
    \label{fig:flow_ecot}
    \vspace{-1mm}
\end{figure}

So far we formulate reasoning and action generation at each time step as batched requests that can be processed in parallel. However, reasoning traces typically span hundreds of tokens, while action decoding involves only a few (around $7$) tokens. In a synchronised setup, this mismatch causes unnecessary latency: the agent must wait for all reasoning steps to complete before it can act (see \cref{fig:flow_ecot}).

\begin{figure}[ht]
    \centering
    \includegraphics[width=\linewidth,trim=1.5cm 7cm 2cm 4.7cm, clip]{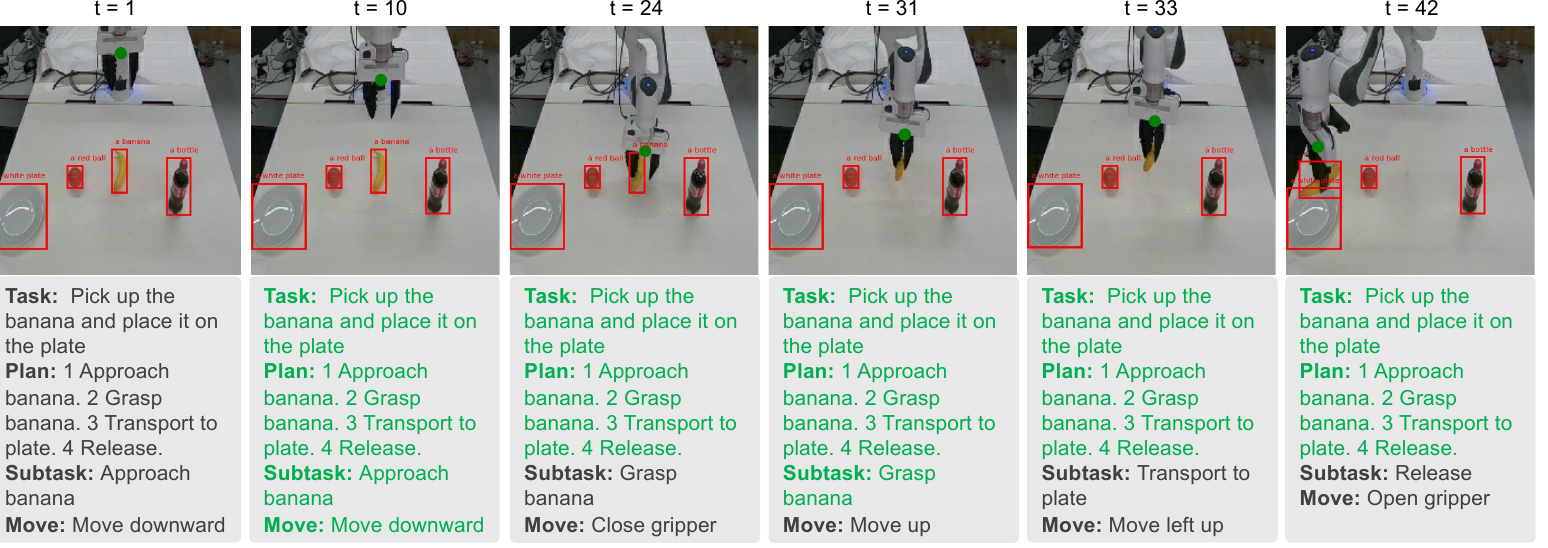}
    \caption{Generated robot rollouts at successive time steps \textbf{(top row)} alongside its reasoning \textbf{(bottom row)}. Between frames, a large part of the reasoning remains unchanged \textbf{(Green)}. At each timestep (t=1, 10, 24, 31, 33, 42), the Subtask updates intermittently, and the low-level Move command adapts continuously as it picks up the banana and places it on the plate.}
    \label{fig:rollouts}
\end{figure}

While ECoT couples reasoning and action for interpretability and causal grounding, this coupling need not be \emph{time-synchronous}. Empirically, high-level elements (e.g., \textsc{Task}, \textsc{Plan}) change slowly across steps; their influence on actions persists even with infrequent updates—see rollouts in \cref{fig:rollouts}, where reasoning traces and visible objects remain stable as actions evolve. Inspired by VLA systems that separate a fast action module from a slower VLM (SmolVLA~\cite{smolvla}; GR00T~\cite{gr00t}), we adopt an asynchronous split: the controller decodes actions from the current observation \(O^t\) and a cached high-level reasoning \(R^c\), while \(R^c\) is refreshed in the background. As shown in \cref{fig:flow_ecot}, asynchronous reasoning action generation effectively reduces stalls and increases action throughput without sacrificing interpretability. Full pseudocode of the above is shown in \cref{alg:ecot_async_batched}.


\begin{algorithm}[ht]
\caption{Asynchronous Parallel Embodied Chain‑of‑Thought}
\label{alg:ecot_async_batched}
\begin{algorithmic}[1]
\Require Time step $t$, Instruction $I^t$, Observation $O^t$, History Reasoning $R^{c}$
\State $c^t \gets \mathrm{encode}(I^t, O^t)$
\For{$i = 1 \,\mathbf{to}\, N+1$ \textbf{concurrently}} 
  \State Lock $R^c$ and autoregressively sample $r_i^t \sim P\bigl(r_i^t \mid c^t, R^{c}[:i]\bigr)$
  \If{$i=N+1$}
  \State $A^t=r_i^t$ 
  \Else 
  \State Lock $R^c$ and update $R^c$ with $r_i^t$
  \EndIf
\EndFor
\State \textbf{wait} until $A^t$ is finished 
\State $A^t \gets \mathrm{decode}(A^t)$
\State \Return $R^c,A^t$
\end{algorithmic}
\end{algorithm}

%% file: sections/experiments.tex
\section{RESULTS}
\label{sec:result}

In this section, we conduct experiments to evaluate the effectiveness of \sysname. Our evaluation focuses on the following key questions: (1) To what extent does parallel reasoning generation improve computational efficiency? (2) Does the proposed method preserve task performance comparable to the sequential baseline? (3) What is the impact of reasoning step parallelization on the overall quality of reasoning?

\subsection{LIBERO Experiments}
\label{sec:libero_setup}

\textbf{Task Setup.} 
We conduct experiments using a Franka Emika Panda robotic arm within the LIBERO environment~\cite{liu2023libero}, a widely adopted benchmark for evaluating generalizable robotic policies. To comprehensively assess policy generalisation, we select four diverse task suites—LIBERO-Spatial, LIBERO-Object, LIBERO-Goal, and LIBERO-Long—each targeting distinct challenges, including spatial configuration, object manipulation, goal specification, and long-horizon task execution. 

\textbf{Training Data and Training Recipe.} 
ECoT models require fine-tuning on each task for optimal performance~\cite{kim2024openvla,zawalski2024robotic}. Training data is generated following the ECoT pipeline~\cite{zawalski2024robotic}, with modifications for reproduction: we integrate Janus~\cite{chen2025janus} for automated scene description and Deepseek-Reasoner~\cite{deepseekai2025deepseekr1incentivizingreasoningcapability} for generating CoT reasoning trajectories from successful demonstrations. We initialise ECoT models from the open-sourced checkpoint~\cite{zawalski2024robotic}, which was pre-trained on Bridge V2 \cite{walke2023bridgedata} and OXE dataset \cite{embodimentcollaboration2024embodimentroboticlearning}. Then we apply LoRA \cite{hu2021loralowrankadaptationlarge} with rank~32 and train for $200,000$ gradient steps using a batch size of~1 distributed across 4 NVIDIA A6000 GPUs. 

\textbf{Baselines.} We compare our proposed method, \sysname, against four baselines: ECoT, the original ECoT model that autoregressively generates the full reasoning chain; ECoT (5-step), a variant that updates low-level reasoning at every timestep but updates high-level reasoning only every 5 timesteps; ECoT (async), a variant originally designed to use two GPUs to asynchronously compute high-level and low-level reasoning, which we adapt to run entirely on a single GPU for fair comparison; and ECoT (Quant), a post-training quantized version of ECoT utilizing Huggingface's Bitsandbytes \cite{bitsandbytes}, selected due to its best acceleration performance compared to other quantization approaches tested \cite{lin2024awqactivationawareweightquantization, xiao2023smoothquant}. For fairness, we exclude methods requiring additional training or architectural changes, such as speculative decoding~\cite{leviathan2023fastinferencetransformersspeculative, kim2023speculativedecodingbiglittle}, and action chunking~\cite{kim2025finetuning, zhao2023learningfinegrainedbimanualmanipulation}.


\label{sec:libero_simulation_results}
\begin{table*}[ht]
\centering
\scalebox{1}{%
\begin{tabular}{l|c|c|c|c|c|c}
\toprule
Method 
  & \shortstack{LIBERO-Object \\ SR (\%) $\uparrow$}  
  & \shortstack{LIBERO-Spatial \\ SR (\%) $\uparrow$}  
  & \shortstack{LIBERO-Goal \\ SR (\%) $\uparrow$} 
  & \shortstack{LIBERO-Long \\ SR (\%) $\uparrow$}  
  & \shortstack{Average \\ SR (\%) $\uparrow$} 
  & \shortstack{Latency per \\ Step (ms) $\downarrow$} \\
\midrule
OpenVLA             & \textbf{87} & 83 & 74 & 55 & 75.3  & \textbf{184 $\pm$ 37} \\ 
ECoT                & 77 & 84 & 75 & 57 & 73.3 & 4997 $\pm$ 691 \\ 
ECoT (5-step)       & 79 & 75 & 72 & 56 & 70.5 & 3514 $\pm$ 969 \\
ECoT (Async)        & 70 & 83 & 80 & 47 & 70.0 & 3655 $\pm$ 773 \\
ECoT (Quant)        & 82 & 82 & \textbf{84} & 57 & 76.3 & 2180 $\pm$ 207 \\
\midrule
\sysname            & 83 & \textbf{85} & 83 & \textbf{69} & \textbf{80.0} & 2156 $\pm$ 353 \\
\texttt{Fast ECoT (Async)} & 75 & 83 & 83 & \textbf{69} & 77.5 & 686 $\pm$ 412 \\ 
\bottomrule 
\end{tabular}%
}
\vspace{4pt}
\caption{LIBERO simulation experimental results. SR = Success Rate.}
\label{tab:LIBERO_results}
\end{table*}

\textbf{Inference Speedup.} As shown in the last column of Tab.~\ref{tab:LIBERO_results}, OpenVLA achieves the lowest latency ($184 \pm 37$ ms) since it directly maps observations to actions without generating intermediate reasoning. In contrast, reasoning-based models such as ECoT incur substantially higher costs. Our method substantially narrows this gap: \sysname reduces latency to $2156 \pm 353$ ms, a 2.3$\times$ speedup over ECoT ($4997 \pm 691$ ms), while \texttt{Fast ECoT (Async)} further reduces it to $686 \pm 412$ ms (nearly 7$\times$ faster), all while retaining reasoning capability. This demonstrates that parallel reasoning effectively amortises the overhead of structured reasoning, bringing latency closer to non-reasoning policies.

\begin{figure*}[!t]
\centering

\begin{subfigure}{0.49\textwidth}
  \includegraphics[width=\linewidth]{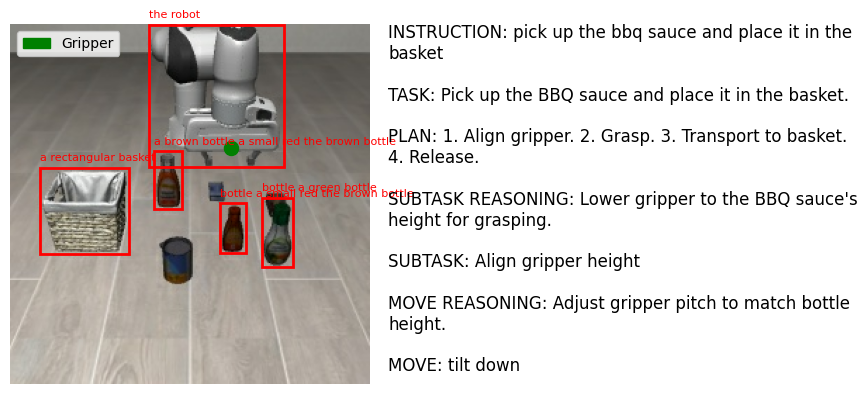}
  \caption{ECoT at $t=143.7$s.}
\end{subfigure}
\hfill
\begin{subfigure}{0.49\textwidth}
  \includegraphics[width=\linewidth]{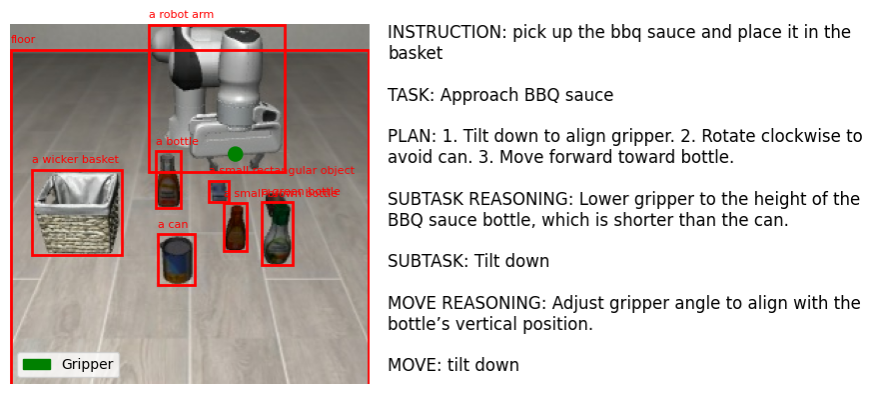}
  \caption{\texttt{Fast ECoT (Async)} at $t=21.4$s.}
\end{subfigure}

\caption{Qualitative examples in LIBERO experiments. \texttt{Fast ECoT (Async)} reaches comparable task states much quicker.}
\label{fig:ecot_sim_examples}
\vspace{-8pt}
\end{figure*}

\textbf{Performance.} Our method consistently improves task performance relative to prior baselines. \sysname achieves the highest average success rate ($80.0\%$), surpassing all ECoT variants and non-reasoning OpenVLA. We attribute this to our parallel reasoning strategy, which smooths temporal inconsistencies and leverages prior reasoning steps. \texttt{Fast ECoT (Async)} remains competitive ($77.5\%$) and achieves the best result on LIBERO-Long ($69\%$), though at some cost in LIBERO-Object due to greater sensitivity to object layouts and temporal mismatch. Interestingly, ECoT (Quant) also improves over vanilla ECoT, likely due to quantisation regularising model behaviour~\cite{krishnan2022quarl}. \cref{fig:ecot_sim_examples} shows that both ECoT and \texttt{Fast ECoT (Async)} produce coherent, grounded reasoning, but the latter reaches comparable states much quicker.

\subsection{Real-world Experiments}
\textbf{Setup.} We validate \sysname on a physical Franka Emika Panda robotic arm equipped with a RealSense D455 camera, providing third-person RGB-D observations. 
We design six manipulation tasks representative of common household scenarios. The evaluation includes both in-distribution and out-of-distribution tasks featuring unseen objects and instructions.
We collect 50 expert demonstrations via Droid teleoperation pipeline~\cite{khazatsky2025droidlargescaleinthewildrobot} to fine-tune ECoT models. We follow the same training data generation process, training recipe, and baselines used in \cref{sec:libero_setup}.



\label{sec:real-world_results}

\begin{table*}[ht!]
\centering
\scalebox{1.0}{%
\begin{tabular}{l|c|c|c|c|c}
\toprule
Method
  & \shortstack{ID \\ SR (\%) $\uparrow$} 
  & \shortstack{OOD Objects \\ SR (\%) $\uparrow$} 
  & \shortstack{OOD Instruction \\ SR (\%) $\uparrow$} 
  & \shortstack{Average \\ SR (\%) $\uparrow$}
  & \shortstack{Latency per \\ Step (ms) $\downarrow$} \\
\midrule
OpenVLA                   & \textbf{83.3} & 68.3 & 42.7 &  64.7 &  196 $\pm$ 32 \\
ECoT                   & 78.3 & 73.3& 40.0 &  64.0&  5556 $\pm$ 384\\
ECoT (5-step)          & 56.6 & 51.7 & 35.0 & 47.8 & 4327 $\pm$ 619\\
ECoT (Async)           & 76.6 & 70.0 & 41.7 & 62.8 & 4206 $\pm$ 323\\
ECoT (Quant)           & 75.0 & 63.3 & 48.3 & 62.2 & 2437 $\pm$ 171 \\
\midrule
\texttt{Fast ECoT}              & 81.6 & 73.3 & \textbf{50.0} & \textbf{68.3} & 2479 $\pm$ 520 \\
\texttt{Fast ECoT (Async)}      & 78.3 & \textbf{75.0} & 42.7 & 65.3 & \textbf{716 $\pm$ 529} \\
\bottomrule
\end{tabular}%
}
\vspace{4pt}
\caption{Real-world experimental results on selected household tasks. SR = Success Rate. ID = In distribution. OOD = Out of distribution.}
\label{tab:real_world_results}
\end{table*}

\begin{figure*}[ht]
\centering

\begin{subfigure}{0.49\textwidth}
  \includegraphics[width=\linewidth]{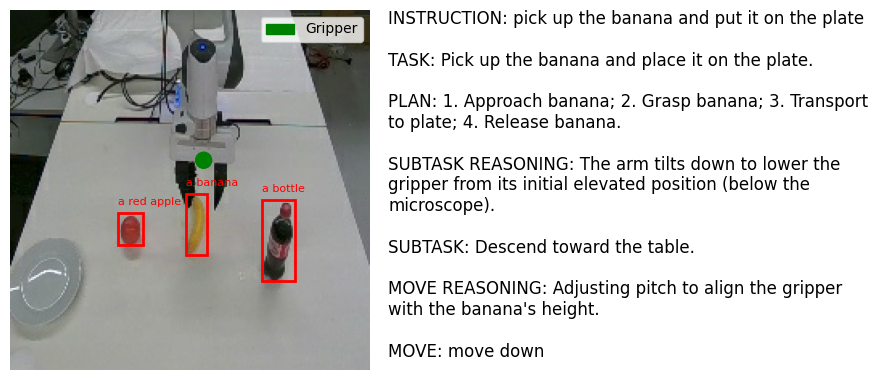}
\caption{ECoT at $t=84.8$s.}
\end{subfigure}
\hfill
\begin{subfigure}{0.49\textwidth}
  \includegraphics[width=\linewidth]{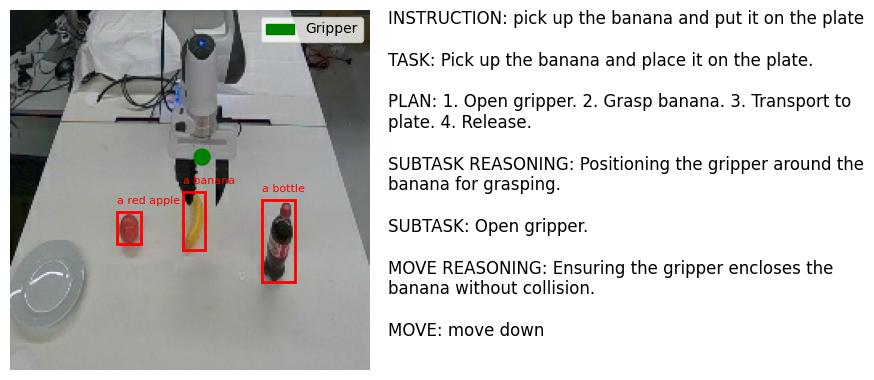}
  \caption{\texttt{Fast ECoT (Async)} at $t=18.7$s.}
\end{subfigure}

\caption{Real-world qualitative examples. \texttt{Fast ECoT (Async)} reaches comparable task states much quicker.}
\label{fig:ecot_real_world_examples}
\end{figure*}

\textbf{Results.}
Tab.~\ref{tab:real_world_results} summarises the performance of our method against baseline approaches on real-world manipulation tasks. 
\texttt{Fast ECoT} achieves the highest overall success rate (68.3\%), with strong improvements in OOD instruction tasks (50.0\%, +10 points over OpenVLA). Meanwhile, the Async version of \sysname yields the best trade-off between performance and efficiency: although its average success rate is slightly lower (65.3\%), it delivers the lowest latency of $716 \pm 529$~ms per step—a $7.7\times$ speedup compared to the original ECoT baseline ($5556 \pm 384$ ms). This highlights the benefit of asynchronous parallel reasoning in real-world deployment, where rapid, low-variance inference is critical. In contrast, prior ECoT variants either sacrifice task performance (ECoT 5-step) or still suffer from high inference latency (ECoT and ECoT-Async), making them less practical for on-robot use. Consistent with the simulation examples, Fig.~\ref{fig:ecot_real_world_examples} shows that both methods produce coherent, grounded reasoning; again, the asynchronous variant reaches comparable task states much quicker. 

\subsection{AutoEval Real-world Experiments}
\textbf{Setup.} We use AutoEval~\cite{autoeval}, an online platform for standardised real-robot policy evaluation. We choose AutoEval both for reproducible, third-party benchmarking and because its tasks/environments are drawn from BridgeData~V2~\cite{walke2023bridgedata}—the same distribution on which ECoT~\cite{zawalski2024robotic} was trained. This lets us run the \emph{vanilla} ECoT \emph{without any finetuning} and directly apply our accelerations (\sysname\ and \texttt{Fast ECoT (Async)}) as drop-in replacements. Concretely, we evaluate four tabletop tasks in two accessible environment (\emph{Drawer}, \emph{Sink}) with 10 trials per task.

\begin{table}[h]
  \centering
  \renewcommand\arraystretch{1.1}
  \setlength\tabcolsep{4pt} 
  \vspace{-2pt}
  \scriptsize 
  \begin{tabular}{@{} l c c c @{}}
    \toprule
    Method & Drawer SR (\%) $\uparrow$ & Sink SR (\%) $\uparrow$ & Latency (ms) $\downarrow$ \\ \midrule
    ECoT                       & \textbf{30} & 30 & 4030 $\pm$ 270 \\
    \sysname                   & \textbf{30} & \textbf{35} & 2105 $\pm$ 324 \\
    \texttt{Fast ECoT (Async)} & 20 & 25 & \textbf{790 $\pm$ 331} \\
    \bottomrule
  \end{tabular}
  \vspace{2pt}
  \caption{AutoEval results on BridgeData~V2 tabletop tasks. SR = Success Rate.}
  \label{tab:autoeval}
  \vspace{-8pt}
\end{table}

\textbf{Results.} \sysname\ and \texttt{Fast ECoT (Async)} achieve performance comparable to ECoT under AutoEval (Tab.~\ref{tab:autoeval}): \sysname\ matches ECoT on \emph{Drawer} (30\%) and slightly exceeds it on \emph{Sink} (35\%), while the asynchronous variant is modestly lower (20–25\%) but offers a favourable accuracy–efficiency trade-off. Note that AutoEval’s environment reset policy introduces distribution shift—e.g., altered object poses/orientations, handle angles, and initial camera/scene configurations—relative to the canonical starts used in the original evaluation. These OOD initialisations make perception and short-horizon control more volatile, reducing absolute SRs across all methods (including ECoT) even though their relative behaviour remains similar.

\vspace{-2.5mm}
\subsection{Updating Frequency of Across-level Reasoning}
Reusing past reasoning steps yields a temporal smoothing effect that dampens noisy fluctuations in planning and stabilises control. We perform a controlled ablation on LIBERO-Object (see Tab.~\ref{tab:ablation}) to examine performance differences resulting from varying the update frequency of high-level and low-level reasoning steps. When \emph{high-level} reasoning is refreshed less frequently (every 5 frames), while \emph{low-level} reasoning remains updated at every step, success rates increase from $77\%$ (baseline with updates at every step) to $79\%$. In \sysname, each reasoning step similarly leverages current visual observations combined with cached reasoning from previous steps, achieving an even higher success rate of $83\%$. 
\textbf{However, excessively infrequent updates reduce performance}. Updating both high-level and low-level reasoning every 5 frames lowers success to $70\%$, and completely infrequent updates (\texttt{$\infty$/$\infty$}) dramatically drop success to $35\%$. Common failures in such scenarios involve delayed task transitions—for instance, after grasping an object, the robot may stall indefinitely rather than moving promptly to the placement step. This underscores the critical balance between beneficial temporal smoothing and necessary action responsiveness.

\begin{table}[h]
  \centering
  \renewcommand\arraystretch{1.1}
  \setlength\tabcolsep{4pt} 
  \vspace{-1pt}
  \scalebox{1.0}{
    \begin{tabular}{@{} l l c @{}}
      \toprule
      High-level step & Low-level step & Success (\%) $\uparrow$ \\ \midrule
      1   & 1   & 77 \\
      5   & 1   & 79 \\
      5   & 2   & 78 \\
      5   & 3   & 75 \\
      5   & 5   & 70 \\
      $\infty$ & $\infty$ & 35 \\
      \sysname & \sysname & \textbf{83} \\
      \bottomrule
    \end{tabular}
  }
\caption{Ablation on update frequency. }
\label{tab:ablation}
\end{table}
Our findings (see \cref{tab:ablation}) align with the results in SmolVLA~\cite{smolvla} and Tab.~2 of the original ECoT work, where the reuse of historical reasoning similarly improved task performance. Thus, moderate reuse of recent reasoning can maintain or even enhance robot performance.
\vspace{0.05em}

\subsection{Action Faithfulness in ECoT Reasoning}
\label{sec:action_faithfulness_in_ecot_reasoning}

\begin{figure}
      \begin{center}
        \includegraphics[width=0.9\linewidth]{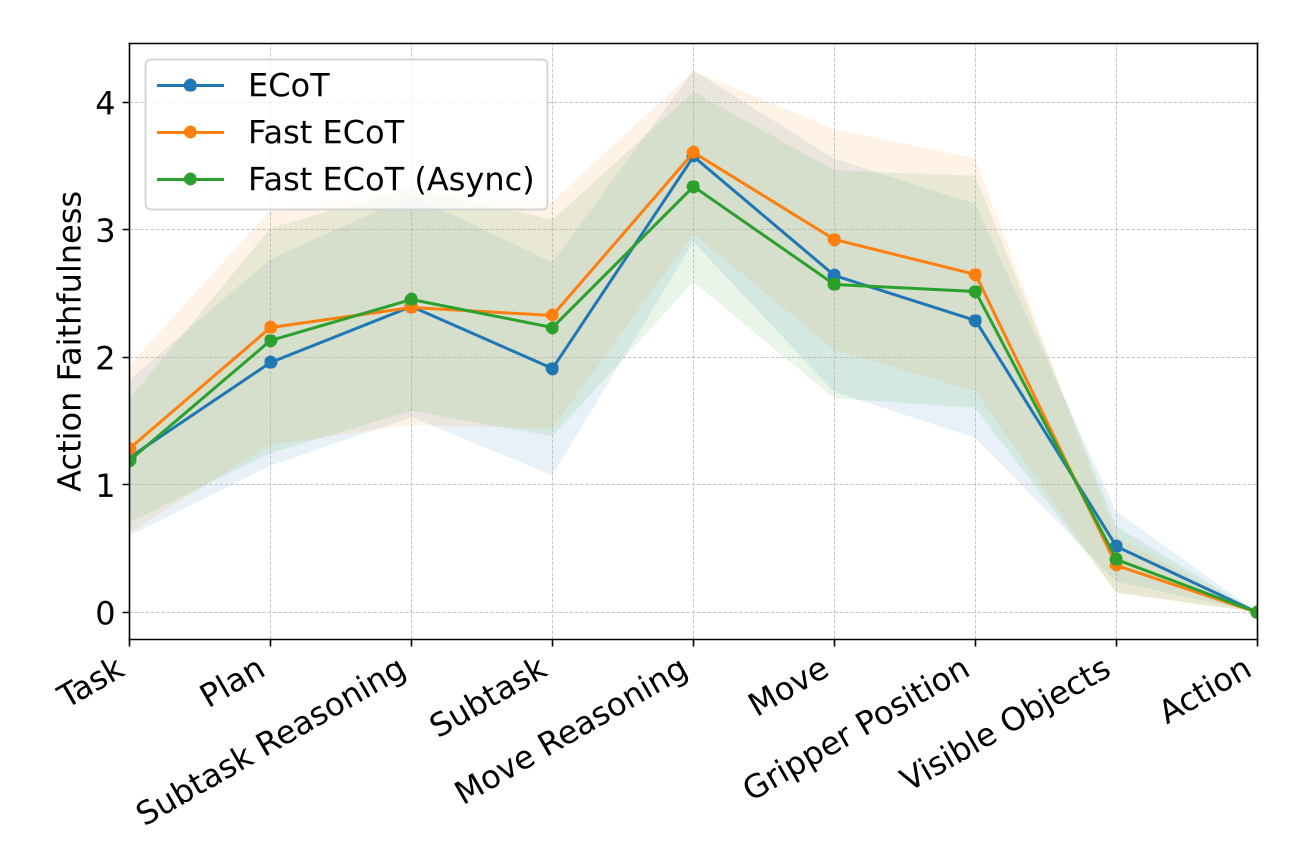}
      \end{center}
      \vspace{-2mm}
      \caption{Action faithfulness (AF) on LIBERO tasks. All graphs are plotted with mean and standard deviation shading across 1000 timesteps.}
    \label{fig:faithfulenss}
    \vspace{-15pt}
\end{figure}

To validate that the generated chain-of-thought (CoT) not only improves performance but also \textit{faithfully explains} the model’s decision-making process, we build on the faithfulness criteria for language models~\cite{Lyu2023FaithfulCR} and introduce a novel quantitative metric - 
\textit{action faithfulness (AF)} - to measure the faithfulness of the CoT reasoning for robotic tasks.  
Formally, given a complete CoT consisting of $N$ intermediate reasoning steps that culminate in a final robot action $A$, we enforce the model to \textit{directly} predict an action $A_i$ after generating only the first $i$ reasoning steps, where $i = 0, 1, \dots, N$. We then compute the L1 distance between $A_i$ and $A$ as the faithfulness score $ AF_i = \|A_i - A\|_1$. Higher L1 distances indicate a greater dependence on subsequent reasoning steps, thereby suggesting higher CoT faithfulness.

We plot action faithfulness scores of \sysname and ECoT for all reasoning steps $\{AF_i\}$  in \cref{fig:faithfulenss}. 
\sysname preserves the faithfulness of the base ECoT model during parallel reasoning generation for most reasoning steps. The action faithfulness without predicting ``Visible Objects" is slightly lower for \sysname, since the high update ratio and one-step delay in visual features might increase its chances to be post-hoc. 
Meanwhile, the asynchronous variant of \sysname generally exhibits a lower faithfulness, probably due to the larger temporal mismatch introduced in async reasoning. 
Faithfulness is notably low when no reasoning steps are generated (i.e., at ``Task"), likely due to fine-tuning from the OpenVLA~\cite{kim2024openvla} checkpoint, which did not require reasoning steps.

%% file: sections/conclusion.tex
\vspace{-5pt}
\section{CONCLUSION}
\label{sec:conclusion}

We present \sysname, an inference-time acceleration method for Embodied Chain-of-Thought (ECoT) reasoning in VLA models. By exploiting structural and temporal locality, \sysname enables (1) reuse of cached high-level reasoning and (2) parallel generation of modular steps. An asynchronous scheduler further decouples reasoning from action decoding, enhancing real-time responsiveness. Requiring no architectural changes or retraining, \sysname integrates easily into existing VLA pipelines. Across simulations and real-world tests, it reduces inference latency by up to 7.5× while maintaining task performance and interpretability, advancing ECoT toward real-time deployment. 